\pdfoutput=1

\documentclass[doc,11pt]{apa7}
\usepackage{mathtools}
\usepackage{amssymb}
\usepackage{newtxtext,newtxmath}  
\usepackage{graphicx}
\usepackage{float}
\usepackage{booktabs}
\usepackage{multirow}
\usepackage{tabularx}
\usepackage{makecell}
\usepackage[font=small]{caption, subcaption}
\usepackage{enumitem}
\usepackage{inconsolata}
\usepackage[autostyle=true]{csquotes}
\usepackage{xparse}
\usepackage{nameref}
\usepackage{hyperref}
\usepackage[natbibapa]{apacite}

\hypersetup{
	colorlinks=true,
	citecolor=blue,
	linkcolor=blue,
	urlcolor=blue
}

\title{Large Language Models provide support for the parallelogram theory of analogy}
\shorttitle{Large Language Models provide support for the theory of Analogy}

\authorsnames[1,1,{1,2},1,1]{Qiawen Ella Liu, Raja Marjieh, Jian-Qiao Zhu,
  Adele E. Goldberg, Thomas L. Griffiths}
\authorsaffiliations{Princeton University, {The University of Hong Kong}}

\authornote{This study was preregistered on AsPredicted
  (\url{https://aspredicted.org/2nk8e9.pdf}). Correspondence concerning this
  article should be addressed to Qiawen Ella Liu, Department of Psychology,
  Princeton University, Princeton, NJ 08540. Email: \texttt{ella.qiawen.liu@princeton.edu}.}

\abstract{Four-term word analogies ($A$:$B$::$C$:$D$) are classically modeled geometrically as parallelograms: adding the vector $B-A+C$ produces $D$. Recent work suggests that this model poorly captures how humans produce analogies, with simple local-similarity heuristics often providing a better account \citep{peterson2020parallelograms}. But does the parallelogram model fail because it is a bad model of analogical relations, or because people are not very good at generating relation-preserving analogies? We compared human and large language model (LLM) analogy completions on the set of problems from \citet{peterson2020parallelograms}. We find that LLM-generated analogies are reliably judged as better than human-generated ones, and are also more consistent with parallelograms in a distributional embedding space. Crucially, we show that the improvement over human analogies is driven by greater parallelogram alignment and reduced reliance on accessible words rather than enhanced sensitivity to local similarity. Finally, fine-tuning GloVe to better satisfy the parallelogram constraint makes the model's top-ranked candidates more likely to be the completions humans and LLMs actually produced, and improves its prediction of human ratings.
Overall, these results provide support for the parallelogram model of word analogies.}

\begin{document}

\maketitle

\section{Introduction}\label{sec:intro}

Analogy has been regarded as a central feature of human creativity and intelligence \citep{gentner1983structure,hofstadter2001analogy,holyoak1995mental}, making investigation of how analogies are formed and why some analogies are better than others a central issue in cognitive science. A simple but canonical paradigm for studying analogy is the four-term word analogy (e.g., \textit{king is to queen as man is to woman}, abbreviated as \textit{king}:\textit{queen}::\textit{man}:\textit{woman}). A classic theory of this type of analogy is the parallelogram model, which posits that concepts exist as points in a geometric mental space where relations are represented as vectors \citep{rumelhart1973model}: to find the word that completes the analogy $A$:$B$::$C$:$?$, one applies the vector difference between $B$ and $A$ to the third term, $C$.

Recent behavioral evidence has, however, called the parallelogram model into question. \citet{peterson2020parallelograms} found that for human-generated word analogies the parallelogram model was outperformed by simple local-similarity heuristics that did not take relational similarity into account (e.g., by choosing a word highly similar to the $C$ term). These findings appear to suggest that geometric models fail to capture how people represent and reason about analogies. Yet an alternative possibility is that people are simply not very good at \emph{producing} analogies. Generating a precise analogy may be cognitively demanding due to time pressure, knowledge constraints, and retrieval failures, which may drive people toward more accessible but less relationally aligned responses, even when they are capable of \emph{recognizing} better analogies when presented with them.

The emergence of large language models (LLMs) offers an opportunity to revisit this question. Despite being highly opaque systems with complex attention mechanisms and inaccessible latent representations, LLMs have shown a surprising capacity to simulate human semantic judgments \citep{piantadosi2024concepts} and demonstrate emergent relational reasoning manifested in tasks like in-context learning \citep{brown2020language}. Crucially, LLMs may not face the same cognitive constraints that may hinder human analogy production. We therefore ask whether LLM-generated analogies are perceived to be higher quality than human-generated ones and whether any such advantage is captured by simple cognitive models such as the parallelogram model or local similarity heuristics. To address these questions, we prompt six state-of-the-art LLMs with a large set of word analogy problems previously given to humans \citep{peterson2020parallelograms} and collect human judgments of how well LLM and human completions preserve the intended relations.

Our results show that LLM analogies are more predictable than people's by both parallelogram \textit{and} local similarity measures. We also find that people reliably judge LLM analogies to be better than human-generated analogies and that the advantage of LLMs over humans is predicted by the use of lower-frequency words and greater parallelogram alignment in an external distributional embedding space. The LLM advantage is driven not by LLMs producing uniformly superior responses, but by humans generating a long tail of low-quality completions. When the comparison is restricted to only the most frequent (modal) responses, the LLM advantage disappears. Preferred analogical completions in this subset of cases, whether generated by LLMs or people, are still predicted by greater parallelogram-alignment and the use of lower-frequency words. Finally, to test whether the parallelogram structure drives more appropriate analogies rather than the effects being due to the particular embedding space we use, we fine-tune the embedding space to better satisfy the parallel structure. Doing so makes the parallelogram model better at recovering the more highly ranked responses from humans and LLMs. Together, these results suggest that while the parallelogram model may poorly describe how humans \emph{generate} analogies, it nonetheless captures the way people \textit{evaluate} good analogies---and that LLMs are better at producing responses that satisfy this structure.

\section{Background}\label{sec:background}

\subsection{Geometric models of analogy}

Early work by \citet{rumelhart1973model} proposed that concepts can be represented as points in a multidimensional space and that relations correspond to difference vectors between those points. Using low-dimensional representations derived from multidimensional scaling, they showed that solving an analogy $A$:$B$::$C$:$?$ (e.g., \textit{rat}:\textit{pig}::\textit{goat}) could be modeled as completing a parallelogram to find a solution at $B-A+C$.

The advent of modern word embeddings has renewed interest in the parallelogram model. Methods such as word2vec \citep{mikolov2013distributed} and GloVe \citep{pennington2014glove} learn vector representations of words from large text corpora, and early demonstrations showed that linear vector arithmetic could recover canonical analogies (e.g., \textit{king} $-$ \textit{man} $+$ \textit{woman} $\approx$ \textit{queen}).
Similar geometric reasoning has also been explored in computer vision, where latent representations support vector-based transformations between visual concepts \citep{reed2015deep,radford2015unsupervised}.

The success of vector arithmetic on word analogies, however, is somewhat fragile. The predicted vector $B-A+C$ often lies closest to one of the input words, so the canonical answer surfaces only because those inputs are excluded from the candidate pool \citep{linzen2016issues}. More generally, canonical responses tend to be predictable from the similarity among $A$, $B$, and $C$ rather than on the relation per se \citep{rogers2017many}. Simple heuristics that rely only on local similarities offer potential alternative ways people may generate analogical completions. \citet{sadler1993context} suggested a \textit{similarity heuristic} that simply ranks responses by their similarity to $C$, and another nearest-neighbor (NN) heuristic that uses the relative similarity of $A$ to $B$ versus $C$ to decide whether to retrieve a response near $B$ or near $C$.

Evaluating the parallelogram rule against both heuristics on a large dataset of human-generated analogies, \citet{peterson2020parallelograms} found that while the parallelogram rule tended to capture the top human responses, similarity-based heuristics provided a better account of the full distribution of responses. As a consequence, they concluded that the parallelogram model may not be a good model of human analogy generation.

\subsection{Analogies by large language models}

A growing body of work suggests that LLMs can solve a range of analogy tasks, from four-term word analogies to matrix reasoning and letter-string analogies, and can support plausible open-ended analogical reasoning when prompted to do so \citep{webb2023emergent,johnson2025large,yasunaga2023large}. \citet{webb2023emergent}, for example, reported that LLMs match or even exceed average human accuracy on standardized multiple-choice analogy tasks (Raven's Progressive Matrices, letter-string analogies, four-term verbal analogies) (but see \citet{lewis2024using} for evidence that these abilities may not generalize beyond familiar task formats). More broadly, LLMs are shown to be good at \textit{in-context learning}: given a few examples, LLMs can infer the latent relationship linking inputs to outputs and apply it to novel cases \citep{brown2020language}. Recent theoretical explanations for in-context learning have grounded these capabilities in neural geometry, demonstrating that task-specific relational information can be extracted as ``function vectors'' from a model's internal activations on a few-shot prompt \citep{todd2023function,hendel2023context}.

Taken together, these findings suggest that LLMs may implement geometric strategies during relational reasoning reminiscent of the parallelogram model proposed by \citet{rumelhart1973model}. Whether such structure resides in the models' own representations, or whether LLMs produce outputs that conform to geometric regularities in the semantic spaces used to evaluate them, has not been investigated. Prior work has also not systematically compared the full distributions of LLM and human responses on a large open-ended generation task, nor evaluated how well existing geometric models account for those distributions. Here, we provide such a comparison and analysis.

\section{Do LLMs Produce Better Analogies than Humans?}\label{sec:better}

We compare LLM and human completions to the same set of analogy problems ($A$:$B$::$C$:$?$), asking whether LLMs produce \emph{different} analogies and then whether the differences are \emph{improvements}. To do so, we generate completions from six LLMs, characterize how their responses differ from humans' in semantic space, and collect human judgments of analogy quality for both populations.

\subsection{Methods}

\subsubsection{Generating LLM analogies}
We prompted six LLMs to complete word analogies, including two
closed-source models (\texttt{GPT-5-mini}~\citep{openai2025gpt5},
\texttt{o4-mini}~\citep{openai2025o3o4mini}) and four open-source models
(\texttt{DeepSeek-V3.1-671B}~\citep{deepseekai2024deepseekv3technicalreport},
\texttt{Qwen3-235B-A22B-Instruct-2507}~\citep{qwen3},
\texttt{Qwen3-32B}~\citep{qwen3},
\texttt{Gemma-3-27B}~\citep{gemmateam2025gemma3}). To minimize artifacts from specific prompt wording, we used four semantically equivalent prompt phrasings for each analogy.\footnote{\small{(1) ``Complete this analogy: A is to B as C is to \_\_\_''; (2) ``What word completes this analogy? A:B::C:?''; (3) ``Analogy: A:B::C:? Answer:''; (4) ``Fill in the blank: A is to B as C is to \_\_\_\_\_''.}} Each was repeated 10 times per analogy, yielding 40 responses per model per analogy. For open models, we used decoding parameters that allow the model to sample with substantial variability rather than always returning a single most-likely answer: \texttt{temperature = 1.0}, which leaves the model's next-token distribution unchanged, and \texttt{top\_p = 0.9}, meaning that at each step the model considers only the most probable candidate words whose probabilities together add up to $90\%$. The two OpenAI reasoning models (\texttt{GPT-5-mini}, \texttt{o4-mini}) were called using the provider's default decoding configuration. We tested LLMs on all 846 analogy items from \citet{peterson2020parallelograms}, drawn from \citet{green2010connecting} (80 analogies), \citet{kmiecik2013semantic} (178 analogies), and \citet{jurgens2012semeval} (588 analogies annotated with the 20 SemEval relations summarized in Table~\ref{tab:semeval_relations}). This yielded 203{,}040 LLM responses comprising 5{,}943 distinct completions; the original \citet{peterson2020parallelograms} dataset contributes 26{,}265 human responses (9{,}136 distinct completions).

\begin{table}[H]
\centering
\small
\caption{Semantic relation categories included from SemEval-2012 Task~2 \citep{jurgens2012semeval}.}
\label{tab:semeval_relations}
\vspace{0.5em}
\begin{tabular}{lll}\hline
\textbf{Category} & \textbf{Subtype 1} & \textbf{Subtype 2} \\
\hline
\textsc{Class-Inclusion} & Taxonomic (\textit{flower}:\textit{tulip}) & Class:Individual (\textit{queen}:\textit{Elizabeth}) \\
\textsc{Part-Whole} & Object:Component (\textit{car}:\textit{engine}) & Collection:Member (\textit{forest}:\textit{tree}) \\
\textsc{Similar} & Synonymy (\textit{car}:\textit{auto}) & Dimensional Similarity (\textit{simmer}:\textit{boil}) \\
\textsc{Contrast} & Contrary (\textit{old}:\textit{young}) & Reverse (\textit{attack}:\textit{defend}) \\
\textsc{Attribute} & Item:Attribute (\textit{beggar}:\textit{poor}) & Object:State (\textit{beggar}:\textit{poverty}) \\
\textsc{Non-Attribute} & Item:Nonattribute (\textit{harmony}:\textit{discordant}) & Object:Nonstate (\textit{laureate}:\textit{dishonor}) \\
\textsc{Case Relations} & Agent:Instrument (\textit{farmer}:\textit{tractor}) & Action:Object (\textit{plow}:\textit{earth}) \\
\textsc{Cause-Purpose} & Cause:Effect (\textit{joke}:\textit{laughter}) & Cause:Compensatory Action (\textit{hunger}:\textit{eat}) \\
\textsc{Space-Time} & Location:Item (\textit{arsenal}:\textit{weapon}) & Time:Associated Item (\textit{retirement}:\textit{pension}) \\
\textsc{Reference} & Sign:Significant (\textit{siren}:\textit{danger}) & Representation (\textit{portrait}:\textit{person}) \\
\hline
\end{tabular}
\end{table}

\subsubsection{Collecting human ratings}
We recruited 390 participants via Prolific (8--10 raters per analogy), all adult native English speakers from the United States who provided informed consent prior to participation in accordance with a protocol approved by the Princeton University Institutional Review Board (protocol \#10859); the study was preregistered on AsPredicted (\href{https://aspredicted.org/2nk8e9.pdf}{link}). We constructed 4{,}048 four-term analogies ($A$:$B$::$C$:$D$) using $A$:$B$::$C$ stems from the SemEval-2012 Task 2 dataset \citep{jurgens2012semeval} and $D$ terms drawn from human responses in \citet{peterson2020parallelograms} and LLM-generated completions to the same items. For both humans and LLMs, we excluded responses provided by only one participant or model, a standard practice in response-generation tasks \citep{nelson2004university}. Stimuli were randomly distributed across 42 lists of about 100 analogies each. Participants rated how similar the relationship between $C$ and $D$ was to the relationship between $A$ and $B$ on a 7-point scale ($1 =$ extremely different, $7 =$ extremely similar). Instructions included examples of similar relationships (reasonable analogies) (\textit{kitten}:\textit{cat}::\textit{chick}:\textit{chicken}) and dissimilar relationships (\textit{chick}:\textit{chicken}::\textit{hen}:\textit{rooster}) (poor analogies). Each list included 5 attention-check trials of obviously dissimilar comparisons that expect ratings $\leq 4$ (e.g., \textit{dog}:\textit{puppy}::\textit{happy}:\textit{sad}); 13\% of participants who failed more than half of these items were excluded from analyses. The remaining participants ranged in age from 18 to 81 years ($M = 44.21$, $SD = 12.33$); 174 identified as female, 157 as male, 3 as non-binary, and 3 preferred not to say.

\begin{figure}[t!]
\centering
\caption{\small{Human--LLM similarity varies across semantic relations. \textbf{Center:} cosine similarity between frequency-weighted centroids of human and LLM analogy completions across the 20 SemEval-2012 semantic relation types. Darker colors indicate greater similarity. \textbf{Insets:} example response distributions for humans and LLMs (bar lengths $=$ proportion of responses). Black bars indicate proportion of each response provided by people.}}
\includegraphics[width=1\textwidth]{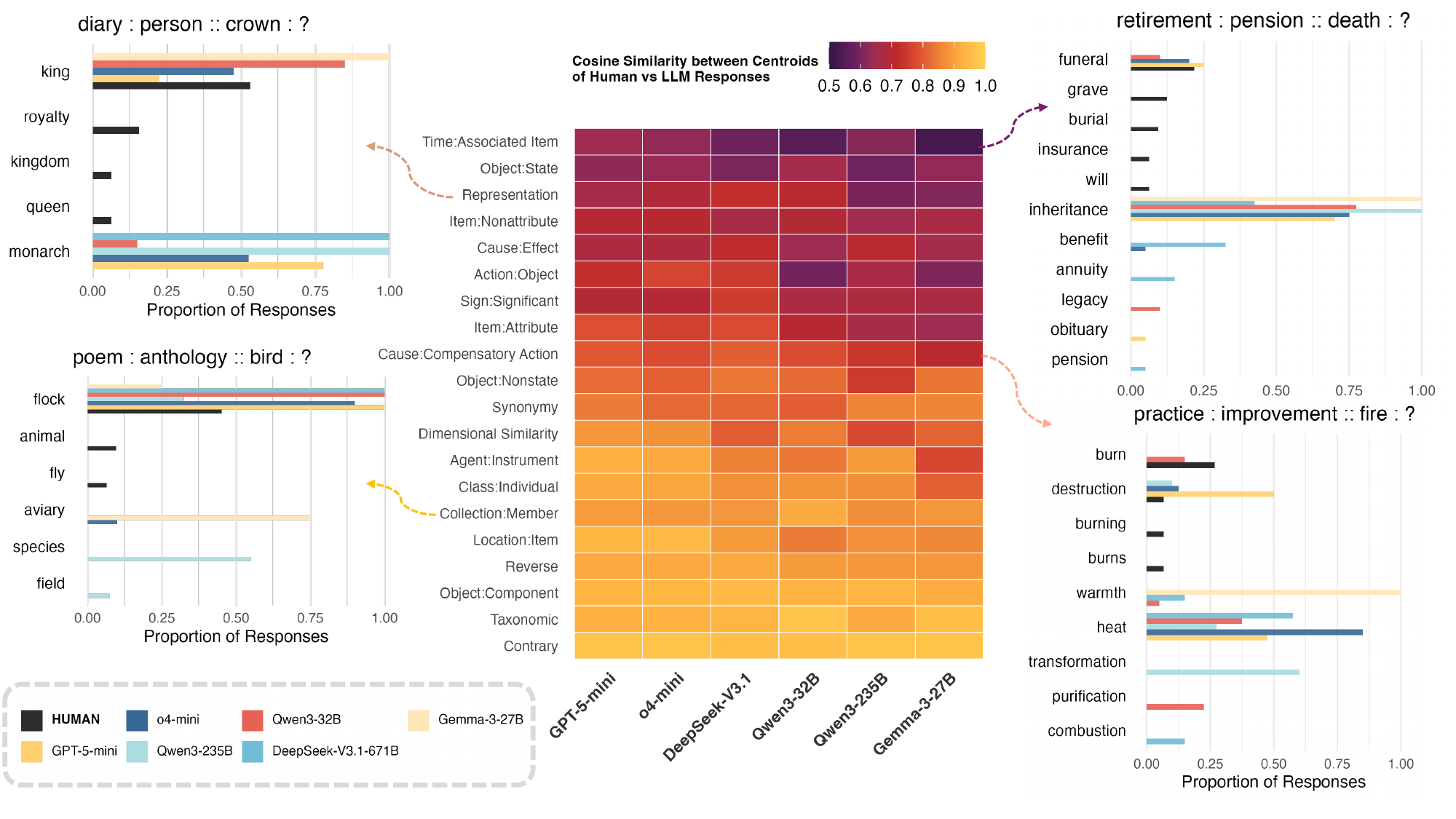}
\label{fig:divergence}
\end{figure}

\subsection{Results}

\subsubsection{LLMs and humans produce different analogies}
To analyze the degree to which LLM responses resemble human responses, we represented words using 300-dimensional GloVe embeddings trained on the 840B-token Common Crawl corpus \citep{pennington2014glove} and compared where their respective responses fall in semantic space. For each analogy stem, we computed a frequency-weighted centroid of responses in embedding space. Let $r$ index unique responses with embeddings $\mathbf{v}_r$ and relative frequencies $f_r$. The centroid is $\displaystyle \mathbf{c} = \sum_r f_r \mathbf{v}_r$. We computed separate centroids for humans and each model and measured their cosine similarity; higher similarity indicates that humans and LLMs generated semantically similar responses, even if the exact words differed (e.g., \textit{cat vs cats}). To contextualize the degree of human-LLM convergence, we also computed human--human convergence by randomly splitting the human responses in half for each stem and measuring the cosine similarity between the centroids of the two halves, repeated 1000 times.\footnote{The centroid is a single-point summary of each response distribution. As a complementary, distribution-level check, we also computed the Jensen-Shannon divergence between the human and LLM distributions per stem (Supplement~A). The two analyses agree closely at the relation level (mean pairwise rank correlation across models $\rho = 0.95$): relations that are most convergent under the centroid measure (e.g., \textit{Contrary} and \textit{Taxonomic}) also show the lowest distributional divergence, while the most divergent relations (e.g., \textit{Time:Associated Item}, \textit{Object:State}, and \textit{Representation}) are the same under both measures.} Across all 846 analogy stems, humans and LLMs produced responses that were broadly similar but reliably distinguishable. The average centroid similarity between randomly split human samples is $.898$ (SE $= .004$ across the 846 stems). Every human--LLM convergence fell reliably below this value ($ps < 10^{-6}$), suggesting that no model's completions resemble human completions as closely as human samples resemble themselves. The two closed-source OpenAI models converged most closely with human completions: \texttt{GPT-5-mini} (mean human--model centroid similarity $= .825$) and \texttt{o4-mini} (mean $= .824$). Among open-source models, \texttt{DeepSeek-V3.1-671B} aligned most closely with humans (mean $= .817$), followed by \texttt{Qwen3-235B} ($.798$), \texttt{Qwen3-32B} ($.797$), and \texttt{Gemma-3-27B} ($.781$). The divergence is also different by relation type (Figure~\ref{fig:divergence}): some relations (e.g., \textit{Contrary}, \textit{Taxonomic}) elicit highly consistent completions across the two populations, whereas others (e.g., \textit{Time:Associated Item}, \textit{Object:State}, \textit{Representation}) diverge sharply.

The divergence between LLMs and humans also suggests that it is unlikely that LLMs were just copying from their training data. Although these classic four-term analogies recur throughout the web text, a majority (58.9\%) of distinct responses generated only by models were never produced by any human given the same stem: e.g., given \textit{beggar}:\textit{poverty}::\textit{child}:?, LLMs most frequently filled in \textit{innocence}, which never appeared in humans' responses (see Supplement~D for more examples of responses provided only by LLMs). Moreover, as shown in the next section, many responses provided only by LLMs were not only valid but rated as \textit{better} than human responses.

\begin{figure}[t!]
\centering
\caption{\small{Relational-similarity ratings for LLM versus human analogy completions. Mean rating differences (LLM $-$ Human) by relation type for six LLMs. Dotted lines show average effects across all relations. *** $p<.001$, ** $p<.01$, * $p<.05$, n.s. $=$ not significant.}}
\includegraphics[width=1\textwidth]{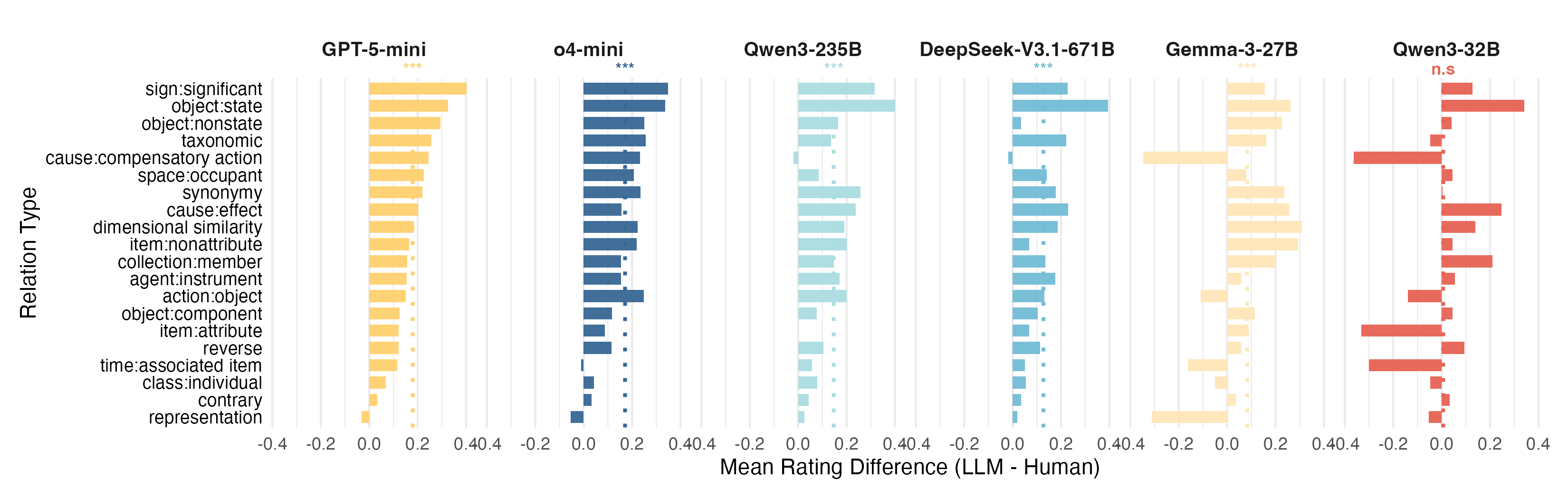}
\label{fig:rating_differences}
\end{figure}

\subsubsection{LLMs produce better-rated analogies}
For each analogy stem ($A$:$B$::$C$), we calculated the mean of the relational-similarity ratings, weighting each completion $D$ by how frequently it was produced (equivalent to averaging ratings as if a completion produced 10 times was rated 10 times). We compared humans to each LLM using paired $t$-tests across all stems.

Figure~\ref{fig:rating_differences} shows the mean rating differences (LLM $-$ Human) for each model, both overall and broken down by the 20 semantic relations. Most LLM analogies were rated statistically significantly higher: \texttt{GPT-5-mini} showed the largest difference with humans ($b = .18$, 95\% CI $[0.14, 0.22]$, $p<.001$), followed by \texttt{o4-mini} ($b = .17$, 95\% CI $[0.13, 0.21]$, $p<.001$), \texttt{Qwen3-235B} ($b = .15$, 95\% CI $[0.11, 0.19]$, $p<.001$), \texttt{DeepSeek-V3.1-671B} ($b = .13$, 95\% CI $[0.08, 0.20]$, $p<.001$), and \texttt{Gemma-3-27B} ($b = .08$, 95\% CI $[0.03, 0.13]$, $p<.001$). The only model whose completions were not rated as significantly better than humans' was \texttt{Qwen3-32B} ($b = .006$, 95\% CI $[-.03, .05]$, $p = .7$).

Beyond receiving higher average ratings, LLMs frequently generated strong completions that no human participant produced. Of the 1{,}869 LLM-only completions for which we obtained ratings, 217 were rated higher than the \emph{best} human completion given the same stem. For example, given \textit{enigma}:\textit{puzzlement}::\textit{explosion}:?, models produced \textit{shock} (rated $6.38$), surpassing the best human response (\textit{chaos}, $5.62$); and for \textit{diary}:\textit{person}::\textit{crown}:?, they produced \textit{monarch}, the more abstract, less frequent, but more relation-preserving term, rated above the best human response (\textit{king}, $4.00$ vs.\ $3.50$) (see Supplement~D for more of such examples).

The magnitude of LLM advantage also varied considerably across relation types. Relations where humans and LLMs already converged showed near-zero rating gaps: \textsc{Contrary} analogies like \textit{happy}:\textit{sad}::\textit{black}:\textit{white} showed near-perfect human--LLM agreement. By contrast, relations with bigger divergence did not always correspond to a bigger rating gap. Some relations (e.g., \textsc{Object:State}) that diverged the most also showed one of the largest rating gaps. For example, given \textit{novice}:\textit{inexperience}::\textit{child}:?, LLMs produced \textit{immaturity} which was significantly better than humans' modal responses \textit{youth} ($b = +.35$, $p<.001$). But some relations e.g., \textsc{Time:Associated Item} ($b = -.05$, $p = .75$) and \textsc{Representation} ($b = -.07$, $p = .41$), which diverged almost to the same extent, showed no significant rating gap. In these cases humans and LLMs drew on different distributions of responses that were nonetheless judged as comparably good, though for different underlying reasons. In \textsc{Representation}, humans and LLMs often read the relation differently, and neither reading was consistently rated as better. The relation between \textit{diary}:\textit{person}, for instance, can be read as a person authors a diary or a diary is about a person. Given \textit{diary}:\textit{person}::\textit{biography}:$?$, humans answered \textit{author} (who writes it) while LLMs answered \textit{subject} (what it is about). The direction reverses on other stems: e.g., given \textit{backdrop}:\textit{vista}::\textit{documentary}:?, LLMs answered \textit{reality} (what a documentary represents) while humans answered \textit{story} or \textit{film} (what documentary is a kind of), and here the LLM reading was rated higher ($5.6$ vs.\ $5.0$ and $3.8$). Because neither system's reading was rated as consistently better, the rating differences roughly offset across the relation's stems, leaving little net advantage for either side. In \textsc{Time:Associated Item}, on the other hand, the two groups drew from a large pool of equally plausible answers. For example, given \textit{infancy}:\textit{cradle}::\textit{childhood}:?, humans answered with the child's sleeping furniture (\textit{bed}, \textit{bunk bed}) while LLMs answered with characteristic play spaces (\textit{playground}, \textit{swing}, \textit{playpen}), which are different objects associated with the same period, all fitting the relation and rated similarly well.

\section{Why are LLM Analogies Better?}\label{sec:exp3}

To understand why LLM-generated analogies receive higher ratings from humans than human-generated analogies, we revisited the word embedding analysis of \citet{peterson2020parallelograms}. Specifically, we ask: (1) How well do parallelogram vs. local similarity heuristics capture the responses produced by humans versus LLMs? (2) What predicts the higher human ratings of LLM-generated analogies? (3) Does increased parallelogram alignment reflect how LLMs internally generate analogies, or does it reflect what humans find to be good analogies?

\subsection{How well do parallelogram vs. local-similarity rules predict human vs. LLM completions?}

\begin{figure}[t!]
\centering
\caption{\small{Performance of humans and six LLMs on three GloVe-based analogy metrics. Bar charts show mean rank with 95\% confidence intervals (lower is better). Insets display cumulative proportion of responses retrieved as a function of rank percentage (log scale).}}
\includegraphics[width=1\textwidth]{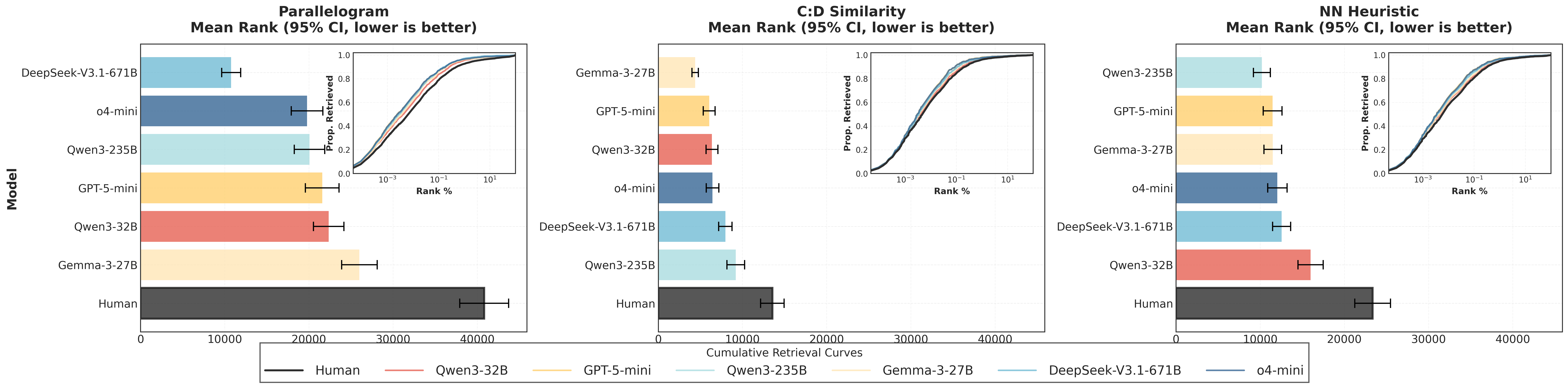}
\label{fig:llm_geometric_models}
\end{figure}

To quantify how well different geometric rules capture human vs. LLM response distributions, we adopted the cumulative proportion retrieved (CPR) analysis from \citet{peterson2020parallelograms}, using the same pretrained GloVe embeddings \citep{pennington2014glove}. We chose GloVe because it explicitly factorizes a word--context co-occurrence matrix, providing a statistically grounded embedding space where geometric properties have distributional interpretations \citep{levy2014neural,ethayarajh2019towards}.

For each analogy stem $A$:$B$::$C$:$?$, each rule induced a ranking over the full vocabulary $V$ ($|V| = 2{,}196{,}015$). We tested three rules:

\begin{enumerate}
    \item \textbf{Parallelogram model.} Predict the completion ($D$ term) by applying to $C$ the same offset that maps $A$ to $B$, i.e., $\widehat{D} = B - A + C$. Then, rank all candidates $w \in V$ by cosine similarity $\cos(w, \widehat{D})$.
    \item \textbf{$C$:$D$ similarity.} Ignore the $A$:$B$ relation and rank candidates only by similarity to $C$, i.e., by $\cos(w, C)$.
    \item \textbf{Nearest-neighbor (NN) heuristic.} First compute whether $A$ is closer to $B$ or $C$. Set the target $T=C$ if $\cos(A, B) > \cos(A, C)$, and $T=B$ otherwise. Then, rank by cosine similarity to $T$, $\cos(w, T)$.
\end{enumerate}

\noindent We then measured, at a rank percentile threshold $\tau$, the proportion of observed responses whose predicted rank falls within the top-$\tau$\% of candidates. Higher CPR at \emph{lower} $\tau$ indicates better prediction.

\subsubsection{All three rules captured LLM responses better than human responses}
On average, the responses of the LLMs were substantially better captured by all three of the rules ($ps<.001$; Figure~\ref{fig:llm_geometric_models}). For example, at the 0.1\% rank threshold, all three rules retrieved a larger proportion of LLM than human responses: parallelogram (86\% vs.~78\%), $C$:$D$ similarity (89\% vs.~84\%), and NN (85\% vs.~80\%).

\subsubsection{Parallelogram underperforms other rules overall, but the gap is smaller for LLMs}
Consistent with \citet{peterson2020parallelograms}, the parallelogram model was relatively good at predicting the top responses, while $C$:$D$ similarity and NN better captured the full distribution. At the same time, the gap between parallelogram and local-similarity rules was notably smaller for LLMs: parallelogram mean ranks were on average 13{,}355 positions higher (worse) than $C$:$D$ similarity for LLMs versus 27{,}264 for humans, and 7{,}799 higher than NN for LLMs versus 17{,}465 for humans. In both cases, the gap between parallelogram and local-similarity heuristic was more than twice as large for humans. As we show in the next section, this apparent advantage of local similarity in capturing \emph{which} completions get produced does not translate into explaining which completions are judged to be \emph{good}: it is parallelogram alignment, not local similarity, that predicts human ratings.

\subsection{What predicts the higher human ratings of LLM-generated analogies?}

To understand what predicts the advantage of LLM-generated analogies, we considered four non-mutually-exclusive predictors: LLMs may receive higher ratings because their responses are (1) more parallelogram-like, (2) more strongly associated with $C$ (i.e., higher local similarity), (3) more aligned with the NN heuristic, or (4) less constrained by lexical accessibility, generating lower-frequency words.

For each completion $D$, we computed: \textbf{parallelogram alignment}, the cosine similarity between $B-A$ and $D-C$; \textbf{$C$:$D$ local similarity}, $\cos(C,D)$; the \textbf{NN heuristic} score selecting either $\cos(C,D)$ or $\cos(B,D)$ depending on whether $A$ is closer to $B$ than to $C$; and \textbf{word frequency}, the log frequency of $D$ via the \texttt{wordfreq} package \citep{robyn_speer_2022_7199437}. We took the weighted average of each metric for human and LLM responses for each stem and computed difference scores (LLM $-$ Human): $\Delta$Parallelogram, $\Delta$$C$:$D$ similarity, $\Delta$NN, and $\Delta \log(\mathrm{freq})$. Because $C$:$D$ similarity and NN are correlated ($r = .66$), these were entered into two multiple regression models (one with $C$:$D$ and one with NN).

\subsubsection{Parallelogram and word frequency explain why LLM analogies are better}
\begin{figure}[t!]
\centering
\caption{\small{Standardized regression coefficients predicting rating differences. *** $p<.001$, * $p<.05$, n.s. $=$ not significant.}}
\includegraphics[width=0.5\textwidth]{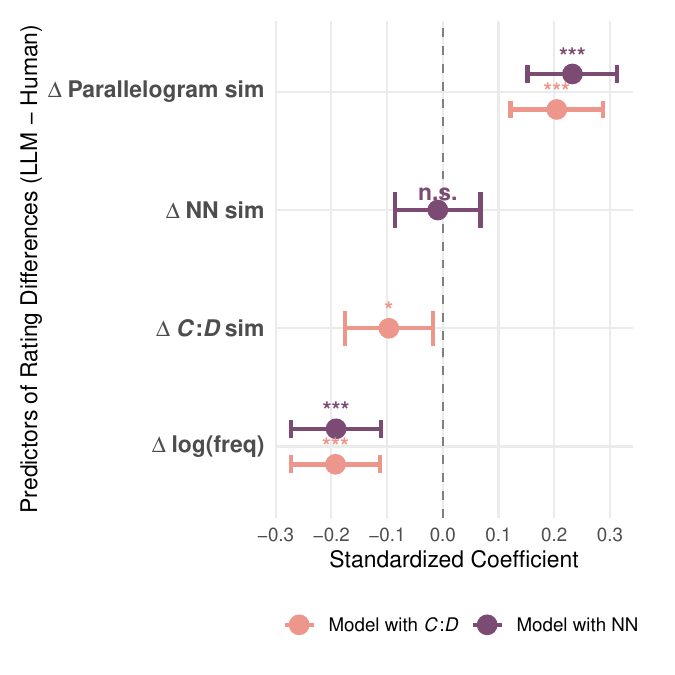}
\label{fig:explanation}
\end{figure}

As shown in Figure~\ref{fig:explanation}, $\Delta$Parallelogram strongly predicted the LLM rating advantage (Model with $C$:$D$: $\beta = 0.205$, $t=4.84$, $p<.001$; Model with NN: $\beta = 0.233$, $t=5.69$, $p<.001$): when LLM completions better satisfied the parallelogram constraint in GloVe, they were also rated as better analogies. Local similarity did not: $\Delta$NN was not a significant predictor ($\beta=-0.009$, $p=.82$) and $\Delta$$C$:$D$ similarity was a significant \emph{negative} predictor ($\beta=-0.097$, $p=.017$). To the extent that LLM completions were more similar to $C$ compared to human completions, the LLM advantage actually shrank. Finally, $\Delta\log(\mathrm{freq})$ was a negative predictor (Model with $C$:$D$: $\beta=-0.192$, $p<.001$; Model with NN: $\beta=-0.191$, $p<.001$), indicating that LLMs outperformed humans more when their completions were less frequent. We also verified that the parallelogram alignment advantage is not a downstream effect for LLMs providing more low-frequency responses, given that the advantage exists at different levels of word frequencies (see Supplement~B for details).

\subsection{Do LLMs' internal representations show the same parallelogram--rating relationship?}
While LLM completions show stronger parallelogram alignment in GloVe space and this alignment predicts human ratings, two interpretations are possible: (1) LLMs excel because their internal representations are more geometrically structured, or (2) parallelogram structure in distributional spaces like GloVe captures what humans find compelling, and LLMs produce better analogies without relying on directly observable geometric representations. To distinguish these, we extracted word representations from the residual stream of each open-source model (\texttt{DeepSeek-V3.1}, \texttt{Qwen3-235B}, \texttt{Qwen3-32B}, \texttt{Gemma-3-27B}), sampling every fourth layer, and computed parallelogram alignment using either isolated word embeddings (e.g., ``king'' presented alone) or contextual embeddings (e.g., ``king'' within \textit{king is to queen as man is to woman}). We then tested how well alignment at each layer predicted human ratings.

\subsubsection{Parallelogram alignment in GloVe predicts human judgments better than LLM internal representations}
Parallelogram alignment in GloVe space strongly predicted human ratings ($\beta = 0.193$, $p < .001$), consistently outperforming all LLM internal representations. Aggregating across layers, LLM effects using isolated embeddings were substantially weaker: \texttt{Gemma-3-27B} ($\beta = 0.094$), \texttt{Qwen3-32B} ($\beta = 0.041$), \texttt{DeepSeek-V3.1} ($\beta = 0.040$), and \texttt{Qwen3-235B} ($\beta = 0.033$); contextual embeddings yielded even weaker predictions (see Supplement~E for details). These results support the second interpretation: LLMs' advantage does not necessarily stem from more parallelogram-like internal representations but from generating analogies that better satisfy relational constraints captured in embedding spaces optimized for semantic representations.

\subsection{The LLM advantage disappears for modal responses, but what makes a good analogy does not}\label{subsec:modal}

While LLMs produce higher-rated analogies than humans on average, this advantage could reflect either that LLMs' responses are overall superior to humans' responses or that LLMs produce fewer low-quality responses. We examined the distributions of responses by humans and LLMs and found human response distributions were considerably more dispersed than LLM distributions: human modal responses accounted for 64\% of total responses, compared to 85\% for LLMs. In other words, humans produced a long tail of less common completions, even though unique completions were removed. LLMs, on the other hand, tended to concentrate their outputs on a few dominant answers. An entropy analysis of the response distributions further shows that human responses were reliably more dispersed than those of every LLM (Supplement~F). If LLMs' advantage stems primarily from humans' long tail of weak responses rather than from LLMs producing better answers across the board, then restricting the comparison to only modal (most frequent) responses should eliminate the LLM advantage.

We tested this prediction by repeating the human--LLM comparison using only modal completions (frequency-weighted, with ties preserved). We found the LLM advantage over humans largely disappeared: no model showed a statistically significant overall advantage over humans ($ps > .05$), while \texttt{Qwen3-32B} and \texttt{Gemma-3-27B} were rated \textit{lower} than human modal responses ($b = -0.10$, $p = .003$; $b = -0.08$, $p = .02$, respectively). All other models' modal responses were statistically indistinguishable from humans' ($ps > .35$). Paralleling this convergence in ratings, most LLMs' modal responses were also not significantly more parallelogram-aligned than humans' in GloVe space (DeepSeek-V3.1-671B: $p < .001$; Qwen3-235B: $p = .03$; all others $ps > .05$).

Crucially, the factors that predicted the LLM advantage in the full dataset remained significant when the analysis was restricted to modal responses. Although the two populations no longer differ on average in parallelogram alignment, alignment still varies from stem to stem, and that variation continues to track which completion is rated better. In our two regression models predicting the rating difference between LLM and human modal completions, $\Delta$Parallelogram was a strong positive predictor (Model with $C$:$D$: $\beta = 0.189$, $p < .001$; Model with NN: $\beta = 0.231$, $p < .001$), and $\Delta\log(\mathrm{freq})$ remained a significant negative predictor (Model with $C$:$D$: $\beta = -0.173$, $p < .001$; Model with NN: $\beta = -0.161$, $p < .001$). $\Delta C$:$D$ similarity was still a significant negative predictor ($\beta = -0.141$, $p < .001$), and $\Delta$NN was a weaker negative predictor ($\beta = -0.086$, $p = .03$) (see Supplement~F for additional visualizations of the modal-response analyses). These results indicate that the mechanisms driving rating differences between human and LLM completions, i.e., parallelogram alignment and lexical accessibility, operate consistently regardless of whether one examines the full distribution or only the dominant responses.

\section{Optimizing an embedding space for parallelogram structure further improves prediction of both human and LLM responses}\label{sec:exp5}

The GloVe embeddings we used were trained only to capture word--context co-occurrences, they were never designed to encode parallelogram structure. If parallelogram structure is what makes an analogy good, then sharpening the space's parallelogram geometry should make it a better account of the completions people and LLMs actually produce. We tested this by fine-tuning the GloVe embeddings to better satisfy the parallelogram constraint and asking whether the resulting space predicts human and LLM completions more precisely than raw GloVe.

\subsection{Approach}
We built a corpus of analogy quadruples ($A$:$B$::$C$:$D$) from pairs of words that share a semantic relation. We drew relations from \citet{bejar1991theories}, who organize them into 81 fine-grained types (e.g., \textit{taxonomic}, \textit{contradictory}, \textit{cause:effect}). For each type, we hand-collected about fifteen word pairs and prompted a language model (\texttt{Claude-Opus-4.8}~\citep{anthropic2026claudeopus48}) to suggest roughly another forty pairs. Hand-picked examples include \textit{flower}:\textit{tulip} and \textit{bird}:\textit{robin} (taxonomic), \textit{alive}:\textit{dead} and \textit{remember}:\textit{forget} (contradictory), and \textit{joke}:\textit{laughter} and \textit{tragedy}:\textit{grief} (cause:effect); the model extended these with pairs such as \textit{spice}:\textit{cinnamon} (taxonomic), \textit{win}:\textit{lose} (contradictory), and \textit{earthquake}:\textit{destruction} (cause:effect).

We then combine these pairs within the same type of relations into four-term analogies: from the taxonomic pairs \textit{flower}:\textit{tulip}, \textit{bird}:\textit{robin}, and \textit{tree}:\textit{oak}, for instance, we form \textit{flower}:\textit{tulip}::\textit{bird}:\textit{robin}, \textit{flower}:\textit{tulip}::\textit{tree}:\textit{oak}, and \textit{bird}:\textit{robin}::\textit{tree}:\textit{oak}. To prevent the model from memorizing the items we evaluate on, we removed any word pair that appeared in a quadruple ($A$:$B$::$C$:$D$) produced by humans or LLMs in our data before recombining. The resulting corpus contains $4{,}422$ unique word pairs, recombined into $123{,}454$ analogies. We then trained a small adapter that learns a single transformation applied to every GloVe word vector \citep{pennington2014glove}, so that the transformed embeddings better satisfy the parallelogram relationship $B - A + C \approx D$. The details of adapter training are given in Supplement~C.

\subsection{Results}
\begin{figure}[t!]
\centering
\caption{\small{Sharpening the parallelogram improves retrieval of human and LLM completions. \textbf{(a)}~Percentage of held-out responses retrieved within the top 1/10/100 neighbors under the parallelogram, $C$:$D$ similarity, and a nearest-neighbor heuristic, before (faded) and after (saturated) fine-tuning. \textbf{(b)}~Cumulative proportion retrieved vs.\ rank for humans (top) and LLMs (bottom, aggregated across six models); solid = after, dashed = before, shading = fine-tuning's gain. Vertical dashed lines mark the rank at which $C$:$D$ similarity overtakes the parallelogram; fine-tuning pushes this crossover much further down the list, extending the range over which the parallelogram remains the best predictor.}}
\includegraphics[width=1\textwidth]{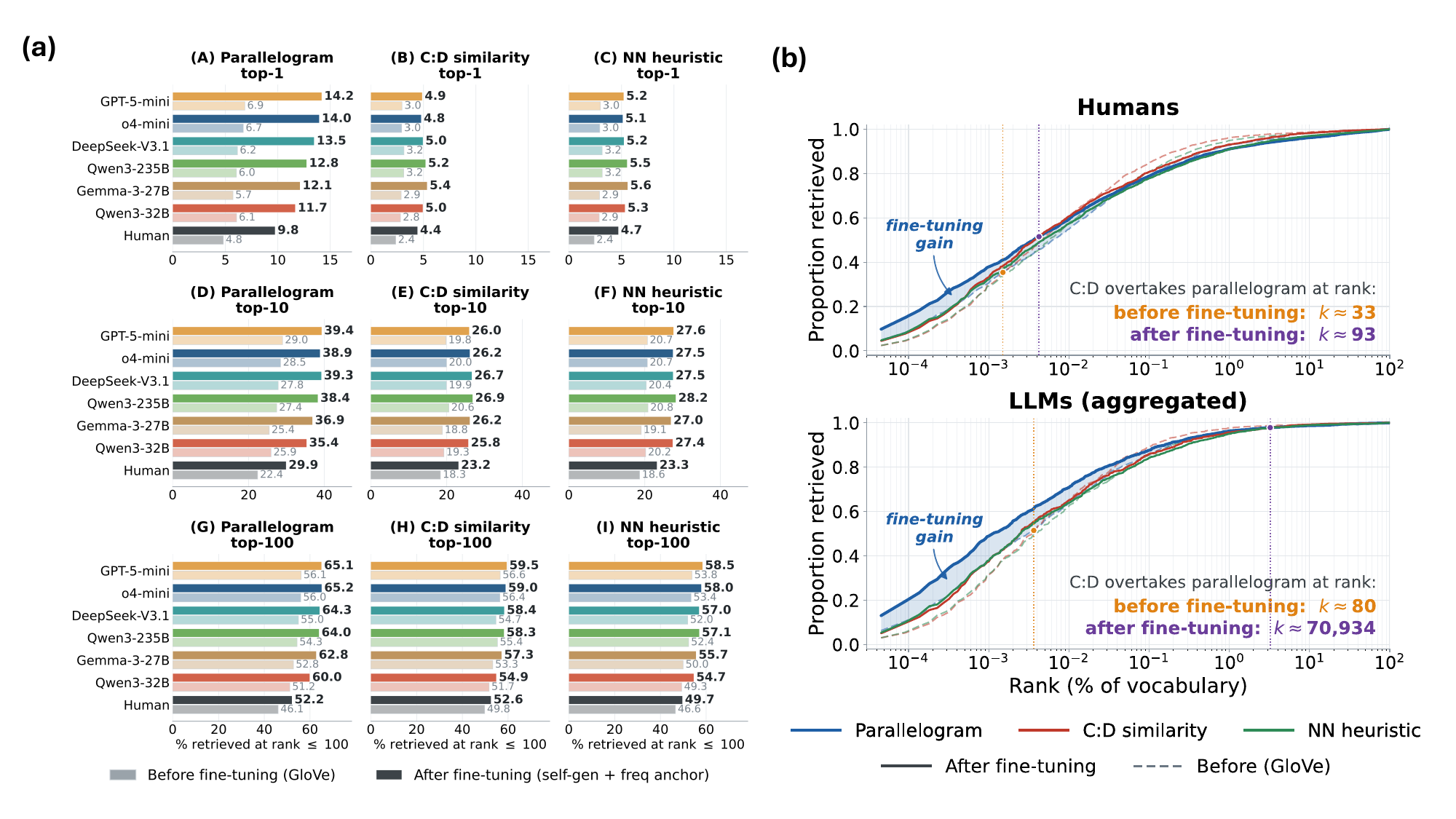}
\label{fig:finetune}
\end{figure}

Strengthening the parallelogram structure of a GloVe space made it markedly better at describing the responses people and models give (see Figure~\ref{fig:finetune}): the top predictions now capture a substantially larger proportion of responses. For human completions, the fine-tuned parallelogram model returned the produced word as its top-1 guess 9.8\% of the time (vs.\ 4.8\% before) and within its top ten 29.9\% of the time (vs.\ 22.4\% before); for the LLMs, on average, top-1 rose from 6.3\% to 13.0\% and top-10 from 27.3\% to 38.0\% (see Figure~\ref{fig:finetune}a for effects broken down by LLMs). $C$:$D$ similarity still outperforms the parallelogram on average rank (mean rank for capturing human responses is 17{,}800 for $C$:$D$ similarity vs.\ 39{,}269 for the parallelogram; LLM mean rank for $C$:$D$ similarity is 9{,}911 vs.\ 12{,}442 for parallelogram). But fine-tuning sharply extended the range over which the parallelogram is the best predictor: for humans, $C$:$D$ similarity did not overtake parallelogram until rank 93 (vs.\ rank 33 before), and for the LLMs the parallelogram led all the way out to rank 70{,}900 before $C$:$D$ similarity caught up (vs.\ rank 80 before fine-tuning; Figure~\ref{fig:finetune}b). The fine-tuned embeddings also predicted human ratings of relational similarity better than the raw embeddings: the correlation between the parallelogram prediction $B-A+C$ and the produced word $D$ rose from Pearson $r = 0.16$ to $0.24$ (Supplement~C).\footnote{This offset-based score is one of two ways to measure the relationship between parallelogram alignment and similarity ratings. The other, used in Section ``Why are LLM Analogies Better'', followed \citealp{peterson2020parallelograms} and measured whether the two relation vectors are parallel, i.e., the cosine between $B-A$ and $D-C$. Because the current adapter is trained to satisfy the offset relation $B-A+C \approx D$ and not the parallel direction score, fine-tuning improves the latter to a lesser degree, from $r = 0.19$ to $0.22$.}

\section{General Discussion}\label{sec:discussion}

Whether simple analogies can be captured via parallelograms in semantic vector spaces is a classic question in cognitive science. By directly comparing LLMs' completions of classic four-term word analogies with human completions, we were able to explore whether today's LLMs solve analogies like people do and, if not, what the differences reveal about the models that describe each system. We found: (1) People judge LLM-generated analogies to be better than human-generated ones, on average. (2) LLM analogies are better predicted by the parallelogram model than human responses, although, much like people's, they are better captured by local similarity heuristics overall. (3) LLM-generated analogies are considered better when they more cleanly instantiate the parallelogram relation and involve less accessible (lower-frequency) words, despite little evidence that LLMs internally represent the parallelogram. (4) This advantage is driven not by LLMs producing superior responses across the board, but by humans generating a long tail of weaker completions. When only modal (most frequent) responses are compared, the LLM advantage disappears, though rating differences between LLM and human modal completions continue to be predicted by greater parallelogram alignment and lower word frequency. (5) Improving the parallelogram structure of a model improves the extent to which parallelogram can capture human and LLMs responses.

Overall, our results suggest that previous failure of the parallelogram model may reflect limitations of human analogy production rather than limitations of the model itself. With LLMs more consistently producing relation-preserving responses, we found the parallelogram model provides a robust account of what makes a word analogy good, even if it may not be a complete account of how either humans or LLMs generate them. In the remainder of the paper, we consider what these results imply for the theories of analogy, why local similarity continues to outperform the parallelogram model in predicting the full distribution, limitations, and directions for future research.

\subsection{Parallelogram captures the perceived quality of analogies}

Compared to humans, LLMs show closer alignment with the parallelogram model's predictions, and this alignment reliably predicts higher human quality ratings. In contrast, local similarity measures such as semantic similarity between $C$ and $D$ do not explain LLMs' advantage and, if anything, predict worse ratings. While \citet{peterson2020parallelograms} suggested that the parallelogram model may simply fail to capture human-like analogies, our findings suggest an alternative interpretation: people may tend to fail to \textit{produce} strong relational analogies, but they nonetheless \textit{prefer} analogies well captured by parallelograms.

One possible interpretation is that LLMs outperform humans because they internally represent analogical relations more geometrically. We found little evidence for this account. The LLMs' own internal representations predict human judgments substantially less well than the parallelogram models' predictions generated in an external embedding space optimized to represent lexical semantics. This suggests that the model that best characterizes analogies need not reflect the process by which those analogies are generated.

We also observe a robust effect of lexical accessibility on relation ratings. LLM completions tend to be lower-frequency than human completions, and people tended to assign lower-frequency completions higher ratings. Crucially, however, the effect of parallelogram alignment remains stable after controlling for accessibility, suggesting that LLM responses are rated better not only because they use less common or more ``clever-sounding'' words, but also because they better satisfy a relational constraint captured by the parallelogram model. Finetuning GloVe to be more parallelogram-aligned provides the strongest evidence that the underlying judgments track relational geometry per se, not GloVe specifically. Together, these findings suggest that the parallelogram model may characterize the structure of good analogies, without implying that either humans or LLMs explicitly encode analogies as parallelograms.

\subsection{Local similarity heuristics and the parallelogram model}

We also replicate and extend a key result from prior work: local similarity heuristics (especially $C$:$D$ similarity) outperformed the parallelogram model as the best overall predictor for both humans and LLMs. \citet{peterson2020parallelograms} attributed this pattern in humans partially to cognitive ease: when rushed or confused, people may be drawn to local similarity judgments rather than engage in full relational reasoning. However, the fact that LLMs exhibit the same preference for local similarity heuristics challenges this account, as LLMs face neither the time pressure nor the cognitive load that constrains human responding. Alternatively, we think local similarity may reliably predict completions because most strong analogies exhibit both relational and semantic similarity between source and target domains. For instance, in \textit{man}:\textit{woman}::\textit{king}:\textit{queen}, \textit{queen} is both relationally appropriate (gender) and semantically similar to \textit{king} (royalty). Even when a clean relational offset is hard to infer due to ambiguity of word meaning or weak relational signal in $A$:$B$, retrieving a word close to $C$ remains a robust fallback for the task.

At the same time, the advantage of local similarity under mean rank should also be interpreted with caution, as it can be disproportionately influenced by a relatively small number of responses that a rule fails to capture at all. Once an observed completion falls far down the ranked list, the difference between 5,000th and 50,000th has little practical consequence for predicting analogy completions, as both are misses. However, the larger value weighs ten times as heavily on the mean. In our case, just 1\% of completions accounts for 50\% of the parallelogram's mean rank for humans, and 80\% for LLMs. Mean rank, therefore, reflects mainly how badly a rule fails on the completions it misses, rather than how well it predicts the completions humans and LLMs are actually likely to produce. Among the highest-ranked candidates, the parallelogram model retrieves a larger proportion of the completions humans and LLMs produced; its disadvantage in mean rank arises farther down the list, where $C$:$D$ similarity places the completions both rules miss less catastrophically. This explains why the gap between parallelogram and $C$:$D$ similarity's mean rank was more than twice as large for humans, as their long tail supplies most of these misses. Our fine-tuning results further support this interpretation, where strengthening the parallelogram structure primarily improved the model's ability to push observed responses toward the top of the ranked list, while having much less impact on how it ordered the large number of unlikely candidates farther down the list.

\subsection{Limitations and Future Directions}

Several limitations of the present work point toward directions for future research. First, the dispersion of LLM responses partly reflects our parameter settings during sampling. With \texttt{top\_p} $=0.9$, the model considers only the most probable candidate words whose probabilities together add up to $90\%$, which cuts off some low-probability tail that humans produce. Our modal-response analyses, which are insensitive to this truncation, suggest the core conclusions do not depend on it, but systematically varying the decoding parameters to match human and LLM dispersion would provide a cleaner test.

Second, the human analogy completions were collected by \citet{peterson2020parallelograms} several years before our study, and a small number of stems have since become outdated, e.g., \textit{everest}:\textit{mountain}::\textit{charles}:\textit{?}, elicited \textit{prince} from human participants at the time of collection, a response that should be updated to \textit{king} now.

A deeper limitation is that human similarity ratings, our primary measure of analogy quality, are themselves shaped by the same accessibility biases that affect production. People tend to rate completions higher when they are stuck on the most salient reading of a relation, even when a less obvious reading is equally or more valid. The relation \textit{diary}:\textit{person}, for example, most naturally calls to mind authorship, i.e., a person writes a diary, so \textit{biography}:\textit{author} gets rated as a better completion than \textit{biography}:\textit{subject}, even though the latter captures the intended representational dimension of the relation just as legitimately. The same tension appears in cases like \textit{elizabeth}:\textit{queen}::\textit{ireland}:\textit{republic}, where \textit{republic} is arguably the more precise answer but loses to more familiar alternatives (\textit{country}) in human ratings; and given \textit{sing}:\textit{dirge}::\textit{cook}:\textit{?}, \textit{feast} preserves the structural parallel more cleanly than \textit{meal} (because a dirge is a genre of song, a feast is a genre of meal), yet raters preferred the simpler answer. In short, our quality measure inherits some of the same biases we are trying to study, which means we likely underestimate how often LLMs are ``right'' in a deeper sense.

This points to another interesting open question. Humans and LLMs do not just differ in how well they complete analogies---they seem to differ in how they interpret the relations underlying a given pair $A$:$B$ in the beginning. When a relation supports multiple valid interpretations, humans and LLMs seem to diverge in reliable ways. Whether this reflects differences in how the two systems represent relations, or simply differences in what gets reinforced during training versus development, is not yet clear. Understanding what the qualitative difference reveals about the underlying representations on each side is a rich direction for future work.

\subsection{Conclusion}

We revisit and lend renewed support to a classic cognitive model of analogy. While the parallelogram model doesn't fully explain how humans or LLMs generate analogies, it reliably predicts what people judge to be good analogies. Moreover, when the geometric structure of an embedding space is explicitly optimized to better satisfy the parallelogram relation, the resulting model becomes better at recovering the completions that humans and LLMs produced and better at predicting human similarity judgments.

Our results also clarify the source of the apparent LLM advantage. LLMs do not outperform humans because they consistently produce superior responses. When only the most frequent responses are compared, the advantage largely disappears. Instead, the difference is driven by humans producing a broader distribution with a long tail of weaker completions, whereas LLMs concentrate their responses on a set of stronger, more relation-preserving completions. This pattern suggests that earlier evidence against the parallelogram model may have partially reflected the variability of human analogy production rather than the inadequacy of the parallelogram model.

More broadly, our results illustrate how LLMs can serve as useful tools for evaluating cognitive theories. Comparing human and LLM behavior allows theories to be tested against multiple intelligent systems rather than against human behavior alone. In the current case, doing so suggests that evidence previously interpreted as a failure of the parallelogram model may instead reflect limitations of human analogy production. Viewed this way, LLMs provide not just another benchmark for analogy performance, but a new source of evidence for distinguishing weaknesses of a cognitive theory from weaknesses of the human behavior traditionally used to evaluate it.

\section*{Data Availability}
All data and analysis code are available on the Open Science Framework at \url{https://osf.io/hqtw2/overview?view_only=cbb155437f03419cb47404574654d90d}.

\section*{Acknowledgments}
This research was supported by Toyota Motor North America, Inc. We also thank Dedre Gentner for thoughtful feedback.

\bibliographystyle{apacite}
\bibliography{parallelogram}

\end{document}